# Multiscale Self Attentive Convolutions for Vision and Language Modeling

**Oren Barkan**

Microsoft

## Abstract

Self attention mechanisms have become a key building block in many state-of-the-art language understanding models. In this paper, we show that the self attention operator can be formulated in terms of $1 \times 1$ convolution operations. Following this observation, we propose several novel operators: First, we introduce a 2D version of self attention that is applicable for 2D signals such as images. Second, we present the 1D and 2D Self Attentive Convolutions (SAC) operator that generalizes self attention beyond $1 \times 1$ convolutions to $1 \times m$ and $n \times m$ convolutions, respectively. While 1D and 2D self attention operate on individual words and pixels, SAC operates on $m$-grams and image patches, respectively. Third, we present a multiscale version of SAC (MSAC) which analyzes the input by employing multiple SAC operators that vary by filter size, in parallel. Finally, we explain how MSAC can be utilized for vision and language modeling, and further harness MSAC to form a cross attentive image similarity machinery.

## 1 Self Attention Revisited

In this section, we provide a brief overview of self attention from [1] and show it can be formulated in terms of $1 \times 1$ convolutions operations. This observation motivates the operators that are proposed in Sections 2-4.

The self attention mechanism [1] transforms a matrix $X = [x_1, \dots, x_n]$, $X \in \mathbb{R}^{d \times n}$ whose columns are the tokens representation, to a new matrix $Y = [y_1, \dots, y_n] \in \mathbb{R}^{d_o \times n}$ as follows

$$Y = V(X)A(X)^T$$

with

$$A(X) = \mathcal{S}\left(\frac{\alpha(X)}{\sqrt{d_a}}\right),$$

where

$$\alpha(X) = Q(X)^T K(X),$$

and $Q(X) = W_Q X$, $K(X) = W_K X$ and $V(X) = W_V X$ are the query, key and value transformations with $W_Q, W_K \in \mathbb{R}^{d_a \times d}$, $W_V \in \mathbb{R}^{d_o \times d}$, and $\mathcal{S}$ is the column-wise softmax operator. Therefore, $A(X)$ is the attention scores matrix and every column $y_i$ is a convex combination of $V(X)$'s columns.

Let $x \in \mathbb{R}^{N_1 \times ... \times N_m}$ be a tensor. We denote

$$x_{i_1...i_k} \triangleq x[i_1, ..., i_k, :, ..., :] \in \mathbb{R}^{N_{k+1} \times ... \times N_m}$$

as the sub-tensor. For example, if $x \in \mathbb{R}^{N \times M \times d}$ is an image with $d$ channels, then $x_{ij}$ is the $d$-dimensional pixel at location $(i, j)$ in $x$.

**Definition 1**: Let $x \in \mathbb{R}^{N \times M \times d}$ and $h \in \mathbb{R}^{n \times m \times d}$ ($n \leq N, m \leq M$) be a signal and a filter, respectively. The discrete convolution operation $x * h$ produces a new tensor $c \in \mathbb{R}^{N \times M}$ s.t.

$$c_{ij} = \sum_{k=1}^{n} \sum_{l=1}^{m} x_{i-\lceil n/2 \rceil + k, j - \lceil m/2 \rceil + l} h_{kl}^T.$$

We further assume $x_{ab} = \vec{0}$ if $a < 1 \lor a > N \lor b < 1 \lor b > M$. Note that in the literature, Def. 1 is equivalent to the discrete (shifted) cross-correlation, while the discrete convolution has a slightly different formulation. However, in this paper, we use both interchangeably.

**Definition 2**: Let $x \in \mathbb{R}^{N \times M \times d}$ and $H \in \mathbb{R}^{F_1 \times ... \times F_L \times n \times m \times d}$ be a signal and a tensor consisting of $L$ filters, respectively. Then, the operation $x \star H$ produces a new tensor $c \in \mathbb{R}^{N \times M \times F_1 \times ... \times F_L}$ s.t.

$$c_{ijf_1...f_L} = \left(x * H_{f_1...f_L}\right)_{ij}.$$

**Lemma 1**: Let $W \in \mathbb{R}^{d' \times d}$ and $X \in \mathbb{R}^{d \times n}$. Then, $WX$ is equivalent to $x \star H$, where $H \in \mathbb{R}^{d' \times 1 \times 1 \times d}$ and $x \in \mathbb{R}^{1 \times n \times d}$ are reshaped versions of $W$ and $X$, respectively.

**Proof:**

$$(x \star H)_{1jf} = \left(x * H_f\right)_{1j} =$$

$$\sum_{k=1}^{1} \sum_{l=1}^{1} x_{1-\lceil 1/2 \rceil + k, j - \lceil 1/2 \rceil + l} H_{fkl}^T = x_{1j} H_{f11}^T = (WX)_{fj} \blacksquare.$$

By using Lemma 1, $\alpha(X)$ can be expressed in terms of $1 \times 1$ convolutions: First, $X$ is convolved with the filters $W_Q, W_k$ and $W_V$ to obtain the query $Q(X)$, key $K(X)$ and value $V(X)$ representations. Then, $K(X)$ is convolved with the filters $Q(X)^T$ to obtain $\alpha(X)$. Finally, $\alpha(X)$ is inversely scaled by $\sqrt{d_a}$ and normalized with a softmax activation to produce the attention coefficients matrix $A(X)$, which is then multiplied by $V(X)$ to obtain the new representation $Y$. This equivalent self attention formulation establishes the framework for 2D self attention, SAC and MSAC.

## 2 2D Self Attention

In this section, we introduce a generalization of the self attention operator from 1D to 2D. In the original form of self attention from [1], each token is represented by a $d$-dimensional vector $x_i \in \mathbb{R}^d$ that attends every other token $x_j$ in the sentence. In this way, the self attention operation is applied over a single axis (e.g. time axis). We propose a spatial (2D) form of self attention that is suitable for images. In 2D self attention, each pixel is transformed to a new space, attends every other pixel in the image to produce attention coefficients that are then being used to form an attentive pixel representation.

Let $x \in \mathbb{R}^{N \times M \times d}$ be a tensor (image) with $d$ channels. Let $H_Q, H_K \in \mathbb{R}^{d_a \times 1 \times 1 \times d}$, $H_V \in \mathbb{R}^{d_o \times 1 \times 1 \times d}$ be the query, key and value filters, respectively and denote

$$q \triangleq x \star H_Q,$$
$$k \triangleq x \star H_K, \quad (1)$$
$$v \triangleq x \star H_V,$$

Then, each pixel in $q \in \mathbb{R}^{N \times M \times d_a}$ is treated as a filter and $q$ is reshaped to $q' \in \mathbb{R}^{N \times M \times 1 \times 1 \times d_a}$ in order to compute the unnormalized attention scores tensor $\alpha \in \mathbb{R}^{N \times M \times N \times M}$ by

$$\alpha = k \star q', \quad (2)$$

where $\alpha_{rtij} = \left(k * q'_{ij}\right)_{rt}$ is the attention score produced by pixel $q_{ij}$ attending pixel $k_{rt}$.

We further introduce a relative positional bias matrix $b \in \mathbb{R}^{N \times M}$ to leverage the relative distance (measured in pixels) between $q_{ij}$ and $k_{rt}$. Then, the attention coefficients tensor $a \in \mathbb{R}^{N \times M \times N \times M}$ is defined as

$$a_{rtij} = \mathcal{S}\left(\frac{\alpha_{rtij}}{\sqrt{d_a}} + b_{|i-r|,|j-t|}\right). \quad (3)$$

Finally, $y \in \mathbb{R}^{N \times M \times d_o}$ is computed by

$$y_{ij} = a_{::ij} \circ v = \sum_{r=1}^{N} \sum_{t=1}^{M} a_{rtij} v_{rt}. \quad (4)$$

## 2.1 Multi-Head 2D Self Attention

Note that $y$ is the result of a *single-headed* 2D self attention. In practice, it is common to use *multi-headed* self attention [1]. The extension from a single-headed to multi-headed 2D self attention is straightforward. To this end, we compute $C$ different 2D self attention heads, independently, where each head $c$ is associated with its own set of filters $H_Q^c, H_K^c, H_V^c$ and produces a corresponding output $y^c \in \mathbb{R}^{N \times M \times d_o}$ (Eq. (4)). Then, we concatenate $(y^c)_{c=1}^{C}$ to form a tensor $y^* \in \mathbb{R}^{N \times M \times C d_o}$ and reduce its channel dimension back to $d_o$ via a $1 \times 1$ convolution layer $H_Y \in \mathbb{R}^{d_o \times 1 \times 1 \times C d_o}$.

## 3 Self Attentive Convolutions

The Self Attentive Convolutions (SAC) operator generalizes the (multi-headed) 2D self attention operator (Section 2) from $1 \times 1$ to $n \times m$ convolutions. Specifically, we use query, key and value filters $H_Q^c, H_K^c \in \mathbb{R}^{d_a \times n \times m \times d}$ and $H_V^c \in \mathbb{R}^{d_o \times n \times m \times d}$. Then, instead of performing 2D self attention in a pixel level, SAC employs 2D self attention in a $n \times m$ patch level. Therefore, SAC allows each patch in an image to attend other patches and vice versa. Figure 1 presents a schematic illustration of a single-headed $2 \times 2$ SAC operator.

In parallel, we can optionally apply a regular convolution layer $H_R \in \mathbb{R}^{d_o \times n \times m \times d}$ to obtain $y' \in \mathbb{R}^{N \times M \times d_o}$ via $y' = x \star H_R$. Then, we concatenate $y$ and $y'$ over the last (channel) dimension to obtain $y'' \in \mathbb{R}^{N \times M \times 2 d_o}$ and reduce its dimension back to $d_o$ by applying a $1 \times 1$ convolution layer $H_\gamma \in \mathbb{R}^{d_o \times 1 \times 1 \times 2 d_o}$ to obtain the final output $\gamma \in \mathbb{R}^{N \times M \times d_o}$ via $\gamma = y'' \star H_\gamma$. In this way, SAC is further combined with a regular convolutional layer.

It is worth noting that the SAC operator can implement any type of convolution (e.g. strided, dilated, etc.) and be straightforwardly integrated in existing network architectures such as ResNets [2], DenseNets [3], etc.

## 4 Multiscale Self Attentive Convolutions

The Multiscale Self Attentive Convolutions (MSAC) operator is a multiscale version of (multi-headed) SAC. MSAC consists of $L$ independent (multi-headed) SAC operators, each analyzing the input in a different resolution using its own set of filters $H_Q^{l,c}, H_K^{l,c} \in \mathbb{R}^{d_a \times n_l \times m_l \times d}, H_V^{l,c}, H_R^l \in \mathbb{R}^{d_o \times n_l \times m_l \times d}$ and $H_\gamma^l \in \mathbb{R}^{d_o \times 1 \times 1 \times 2d_o}$ $(1 < l < L)$. Then, the outputs $(\gamma^l)_{l=1}^L$ from the SAC components are concatenated over the channel axis to form a supertensor $\psi \in \mathbb{R}^{N \times M \times d_o L}$. Finally, we reduce the channel dimension by using a $1 \times 1$ convolution layer $H_\phi \in \mathbb{R}^{d_o \times 1 \times 1 \times d_o L}$ to obtain the final output $\phi \in \mathbb{R}^{N \times M \times d_o}$ ($\phi = \psi \star H_\phi$) that forms a multiscale self attentive representation of $x$. Figure 2 depicts a schematic illustration of MSAC.

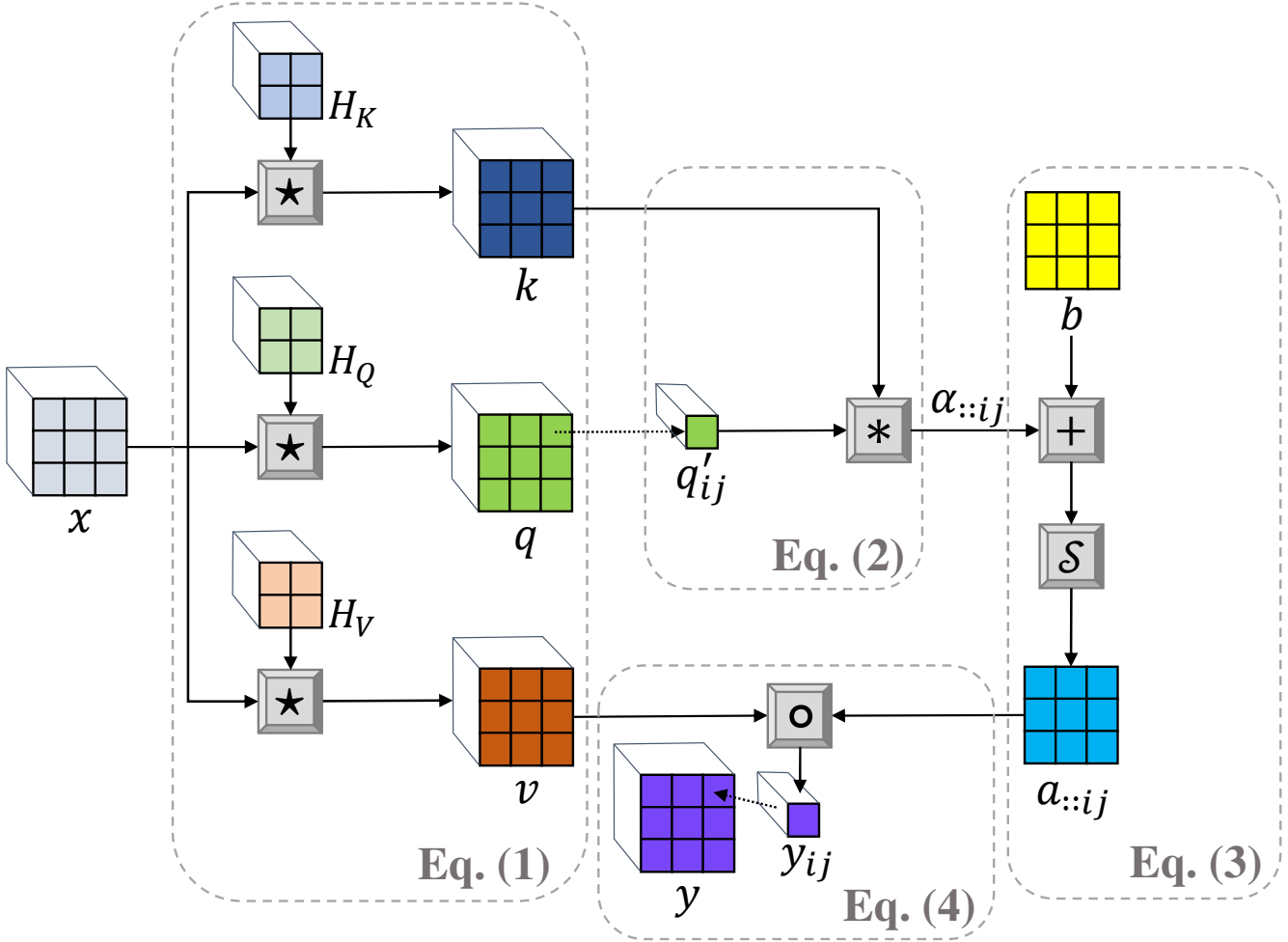

Figure 1. An illustration of a single-head $2 \times 2$ SAC. The optional regular convolution layer $H_R$ is omitted. See Section 3 for details.

## 5 MSAC for Language Modeling

MSAC allows for a multiresolution self attentive analysis of sentences and free text, and can potentially replace the original self attention mechanism from [1] that is widely used in recent language models such as Transformer [1] and BERT [4]. MSAC.

Let $x \in \mathbb{R}^{1 \times M \times d}$ be a sentence of length $M$, where each word is represented by a $d$-dimensional embedding vector. Then, MSAC can be applied to $x$ as explained in Section 4 and depicted in

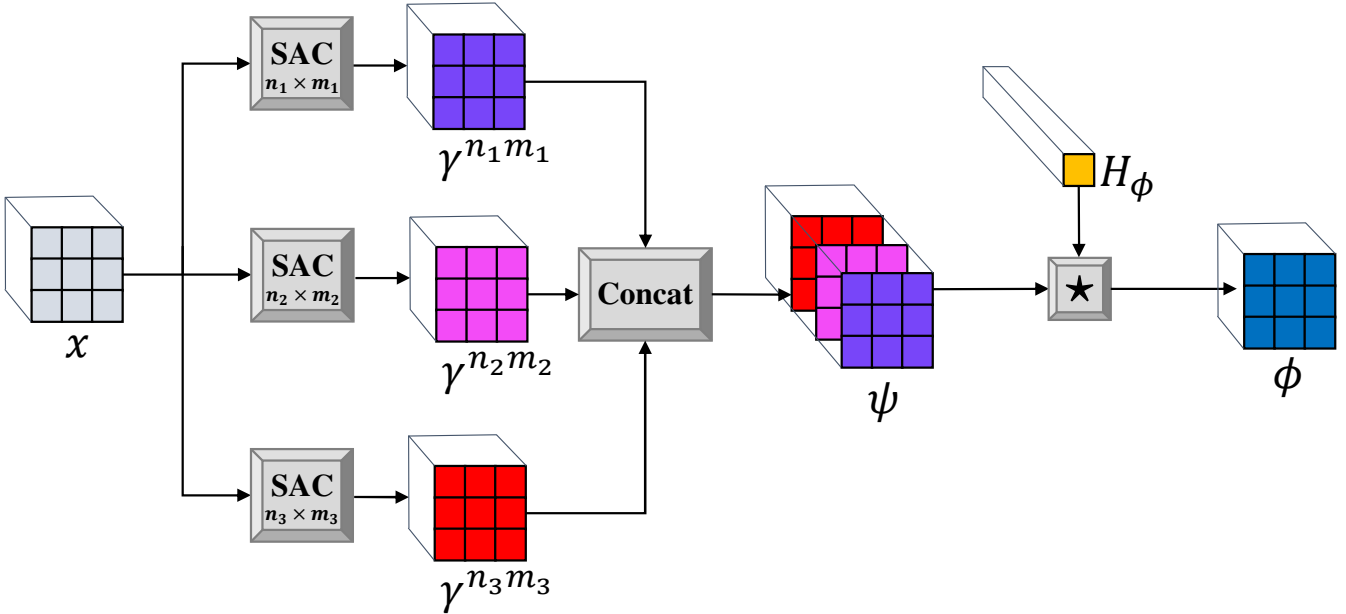

Figure 2. MSAC illustration. The figure depicts a MSAC block consisting of three SAC operators that vary by resolution. See Section 4 for details.

Fig. 2, with $n_l = 1$ $(1 < l < L)$. For example, consider the sentence 'the quick red fox jumps over the lazy brown dog', and assume we use a MSAC operator that combines $1 \times 1$, $1 \times 2$ and $1 \times 3$ SAC operators. Then, besides the $1 \times 1$ SAC operator that is equivalent to the regular self attention and allows each word to attend all other words (e.g. 'fox' attends 'dog' and vice versa), MSAC further allows every pair and triplet of words to attend all other pairs and triplet, respectively. Specifically, the 'red fox' and 'quick red fox' can attend the 'brown dog' and the 'lazy brown dog' by using the $1 \times 2$ and $1 \times 3$ SAC operators, respectively. This unique property might result in a better encoding of semantic information, as it allows the model to capture relations between (parts of) sentences.

## 6 MSAC for Vision Modeling

MSAC can replace the regular convolutional layers that are widely used in many state-of-the-art vision models [2], [3], as explained in Section 4. The multiscale nature of MSAC endows deep models with the ability to focus on different scales at different stages (layers). For example, the first layers in the network can benefit from SAC operators of high scales that learn the attentive relations between patches (e.g. $5 \times 5$ via a cascade of two $3 \times 3$ filter sets). Obviously, using $1 \times 1$ SAC operators in the first layer of the network is probably redundant, as it results in an attentive pixel-pixel analysis that makes less sense. However, in subsequent layers, where the spatial dimensions $N$ and $M$ are already reduced (either by strided convolutions or pooling operations), $1 \times 1$ SAC operators might contribute. Note that the special case of $1 \times 1$ SAC is essentially equivalent to 2D self attention (Section 2).

### 6.1 Cross Attentive Image Similarity

Let $x, z \in \mathbb{R}^{N \times M \times d}$ be images with $d$ channels. MSAC operators enable a mechanism for cross attentive image similarity, in which each patch in image $x$ attends all other patches in image $z$ and vice versa. To this end, we concatenate $x$ and $z$ to form a new image $t = [x, z] \in \mathbb{R}^{N \times 2M \times d}$. Then, a network architecture that employs MSAC is capable of performing cross attentive analysis of the input $t$. For example, if both $x$ and $z$ present two different scenes of a boy playing basketball, then the basketball and the boy in image $x$ can attend the basketball and the boy in image $z$, respectively.

We can optionally use a *segment augmentation* in order to facilitate an explicit distinction between patches of image $x$ and patches of image $z$. A segment augmentation can be implemented by adding learnable tensors $x^{seg} \in \mathbb{R}^{N \times M \times d}$ and $z^{seg} \in \mathbb{R}^{N \times M \times d}$ to $x$ and $z$, respectively. Another option is to augment the channel dimension of $x$ and $z$ with $x^{seg} \in \mathbb{R}^{N \times M \times d_{seg}}$ and $z^{seg} \in \mathbb{R}^{N \times M \times d_{seg}}$ (via a concatenation), respectively.